# Understanding Deep Neural Network Predictions for Medical Imaging Applications

Barath Narayanan Narayanan[a,b*], Manawaduge Supun De Silva[a], Russell C. Hardie[a], Nathan K. Kueterman[a], and Redha Ali[a]

[a]Department of Electrical and Computer Engineering, University of Dayton, 300 College Park, Dayton, OH, 45469

[b]Sensors and Software Systems Division, University of Dayton Research Institute, 1700 South Patterson Blvd., Dayton, OH, 45409

*Barath Narayanan Narayanan, Email: narayananb1@udayton.edu

**Abstract**

Computer-aided detection has been a research area attracting great interest in the past decade. Machine learning algorithms have been utilized extensively for this application as they provide a valuable second opinion to the doctors. Despite several machine learning models being available for medical imaging applications, not many have been implemented in the real-world due to the uninterpretable nature of the decisions made by the network. In this paper, we investigate the results provided by deep neural networks for the detection of malaria, diabetic retinopathy, brain tumor, and tuberculosis in different imaging modalities. We visualize the class activation mappings for all the applications in order to enhance the understanding of these networks. This type of visualization, along with the corresponding network performance metrics, would aid the data science experts in better understanding of their models as well as assisting doctors in their decision-making process.

## 1. Introduction

This paper describes a study that supports the apprehension of the results predicted by deep neural networks applied towards medical imaging analysis. Several machine learning and deep learning architectures have been proposed in the literature for automated Computer-Aided Detection (CAD) tools for various applications [1-4]. In the past few years, deep learning networks have been used widely in medical imaging applications [1, 3]. Residual Networks (ResNet) [5] and GoogLeNet [6] are some of the most popular networks used in this field. The availability of a vast variety of networks raises the question of choosing the optimal network for a given disease/condition. In a data science perspective, optimal results could be measured in terms of overall accuracy, confusion matrix, precision, recall, Receiver Operating Characteristics (ROC) curve, or any other performance metric. However, these optimal results might not be satisfactory for the doctors if the results are not interpretable. Determining the Region of Interest (ROI) that contributed to the decision making of the network will enhance the understanding for both data science experts and clinicians.

In this research, we investigate the results provided by networks for a given image type (microscopic, retinal, Magnetic Resonance Imaging (MRI), and chest radiograph). We visualize the Class Activation Mapping (CAM) [7] results for various medical imaging applications to localize the targeted image regions. The probability predicted by the network for each class gets mapped back to the final convolutional layer of the respective network to highlight the discriminative regions that are specific to each class [7]. The CAM for a specific class is the result of the activation map of the ReLU (Rectified Linear Unit) layer after the final convolutional layer. This is determined by how much each activation contributes to the final score of that particular class. The originality of CAM is the global average pooling layer applied following the last convolutional layer based on the spatial location in order to generate the weights [7]. Therefore, it allows distinguishing the areas within an image that differentiates the class specificity prior to the softmax layer, which leads to the probability predictions [7].

Visualization using CAM for deep neural networks provides more confidence in its predictions to the users. It also allows the expert readers, doctors and clinicians to see the image regions that are deemed to be highly salient in the detection process. We apply this visualization technique for malaria detection on microscopic images, Diabetic Retinopathy (DR) detection on retinal images, brain tumor detection on MRI scans, and finally, tuberculosis detection on chest radiographs.

To maintain the homogeneity across all these applications, we solely study the performance of transfer learning approaches using GoogLeNet and ResNet. Figure 1 presents the top-level block diagram of the transfer learning methodology adopted in this study. We implement these techniques for publicly available datasets thereby setting a new benchmark for each application. Results presented for the publicly available datasets would grant the capability for researchers to replicate and enhance them further. This type of analysis would assist the doctors and data science experts in selecting the optimal model of their choice.

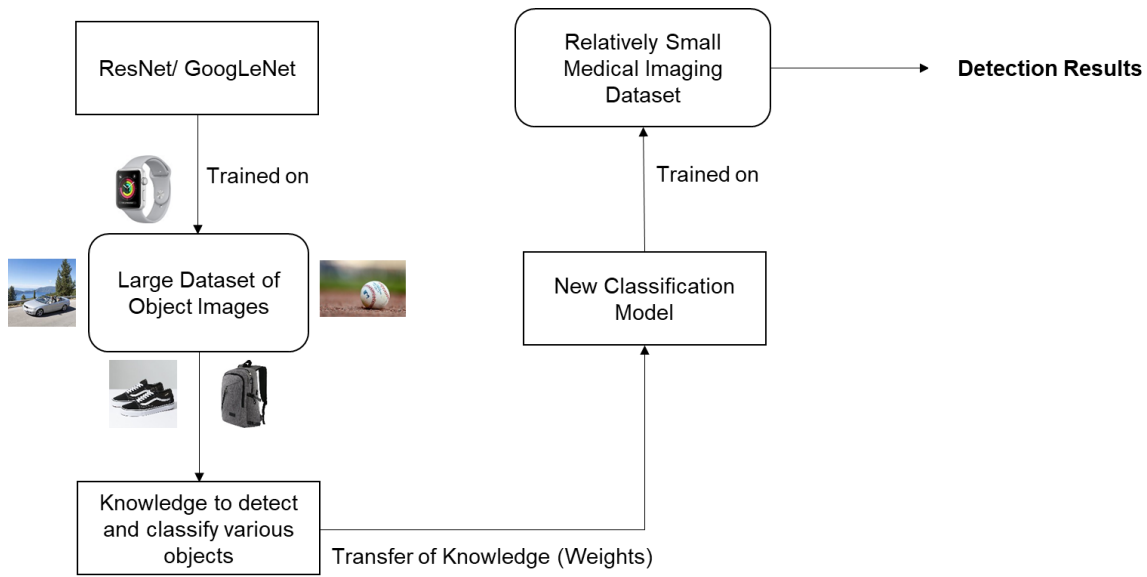

**Figure 1**: Top-level block diagram of the transfer learning approach adopted.

The remainder of this paper is organized as follows. Sections 2-5 present the results obtained for CAD of malaria, DR, brain tumor, and tuberculosis respectively. In each of these sections, we describe the dataset along with the experimental results obtained in terms of both performance metrics and CAM results. Finally, discussions and conclusions are offered in Section 6.

## 2. Malaria Detection

In this section, we study the performance and visualize the CAM results of GoogLeNet and ResNet for the detection of plasmodium on cell images captured using digital microscopy. Plasmodium malaria is a parasitic protozoan that causes malaria in humans.

*2.1. Dataset*

We make use of a publicly available dataset provided by the National Institutes of Health (NIH) for the classification of cell images [8, 9]. Any cell image that contains plasmodium is marked as 'parasitized' by expert analysts and 'uninfected' otherwise. The dataset contains a total of 27,558 images with equal distribution of parasitized and uninfected cells. Figures 2 and 3 present sample images marked as parasitized and uninfected by expert readers.

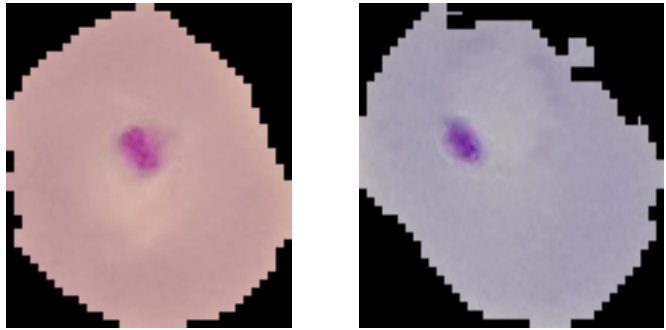

**Figure 2**: Sample cell images marked as 'parasitized' by expert readers.

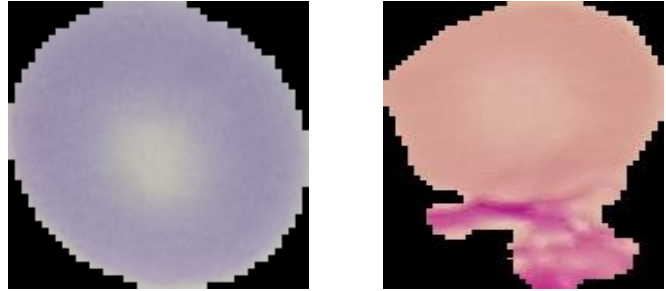

**Figure 3**: Sample cell images marked as 'uninfected' by expert readers.

For this application, we perform a hold-out validation study. We split the dataset into groups of 80% and 20% for training and testing respectively. We utilize a subset of 10% from our training data for validation purpose in order to fine-tune our hyperparameters. Table 1 presents the distribution of the dataset.

**Table 1**: Malaria dataset distribution.

| Type of Dataset | # Parasitized images | # Uninfected images |
|---|---|---|
| Training | 9921 | 9921 |
| Validation | 1102 | 1102 |
| Testing | 2756 | 2756 |

## 2.2. Preprocessing

Images from digital microscopy are captured under different illumination conditions. To aid our classification architecture, we preprocess the images using a color constancy technique [10, 11] to maintain the color contrast across all images. Later, we resize the images to match with the ResNet and GoogLeNet architectures. Figures 4 and 5 present the results obtained using the color constancy preprocessing for different images [10]. The code utilized for color constancy is available at [11].

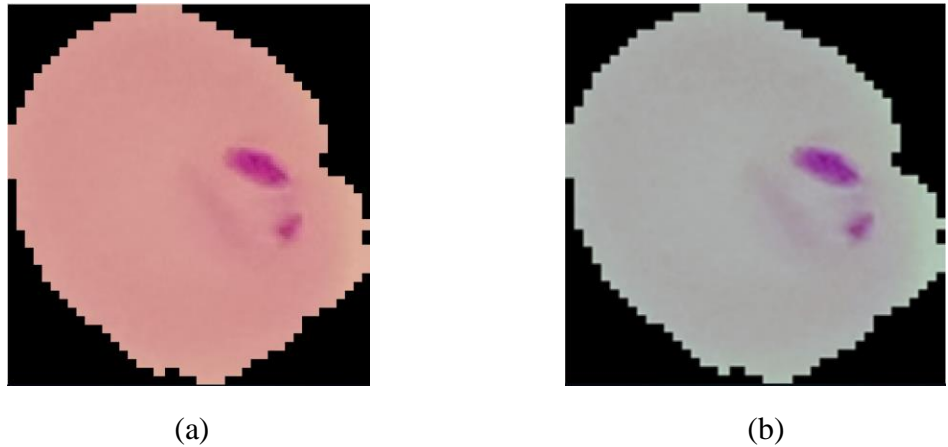

(a)           (b)

**Figure 4**: Preprocessing results: (a) raw image, (b) color constancy and resizing.

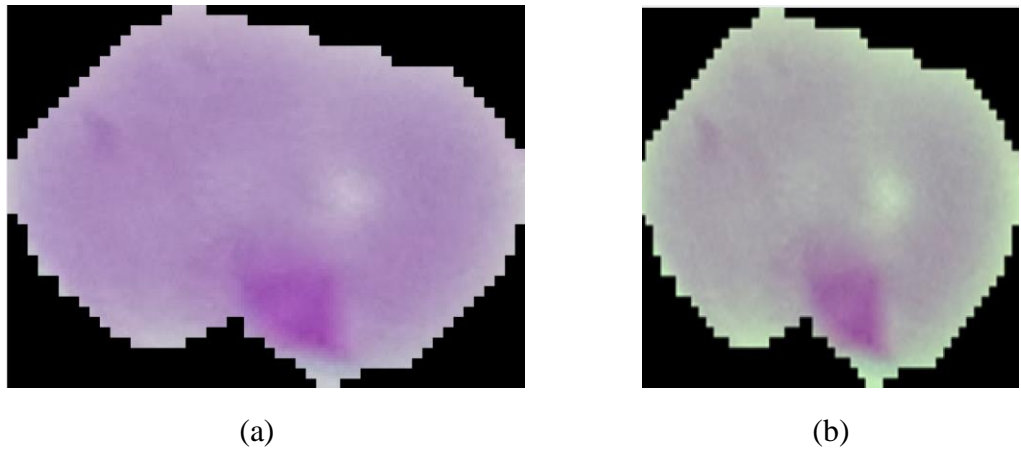

(a)           (b)

**Figure 5**: Preprocessing results: (a) raw image, (b) color constancy and resizing.

## 2.3. Performance Summary

In this section, we study the results obtained using the transfer learning based approaches with ResNet and GoogLeNet. Figures 6 and 7 present the confusion matrices and ROC curves obtained using these approaches for malaria detection. Table 2 presents the performance summary in terms of overall accuracy and Area under ROC (AUC) for malaria detection using these networks.

Table 2: Overall accuracy and AUC for malaria detection.

| Method | Overall Accuracy (%) | AUC |
|---|---|---|
| GoogLeNet | 96.5 | 0.9929 |
| ResNet | 96.6 | 0.9934 |

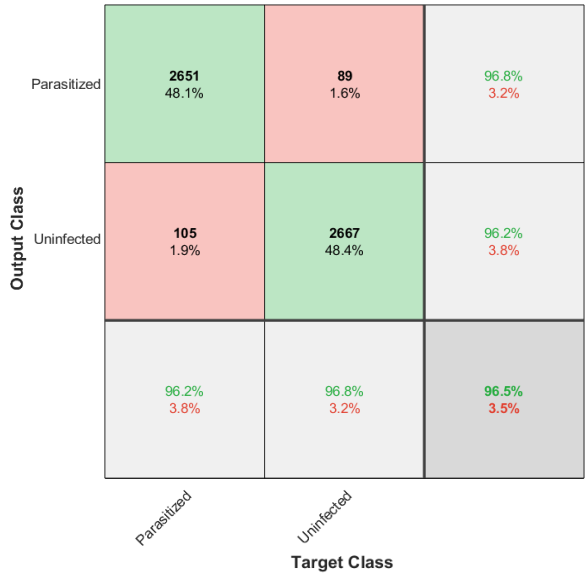
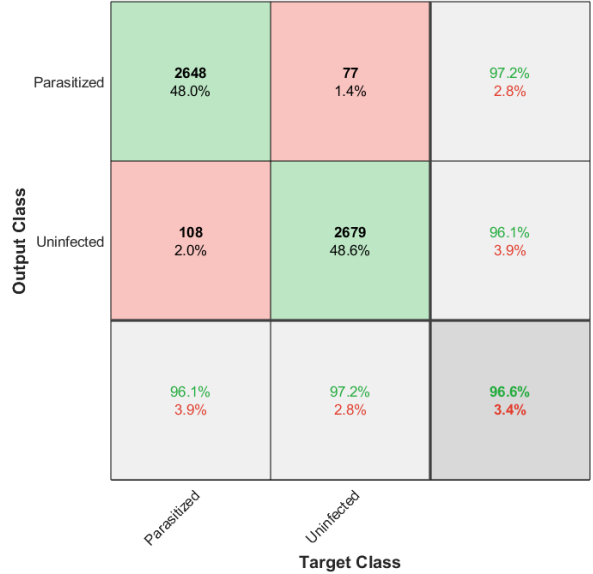

(a)      (b)

**Figure 6:** Confusion matrices obtained for malaria detection: (a) GoogLeNet and (b) ResNet.

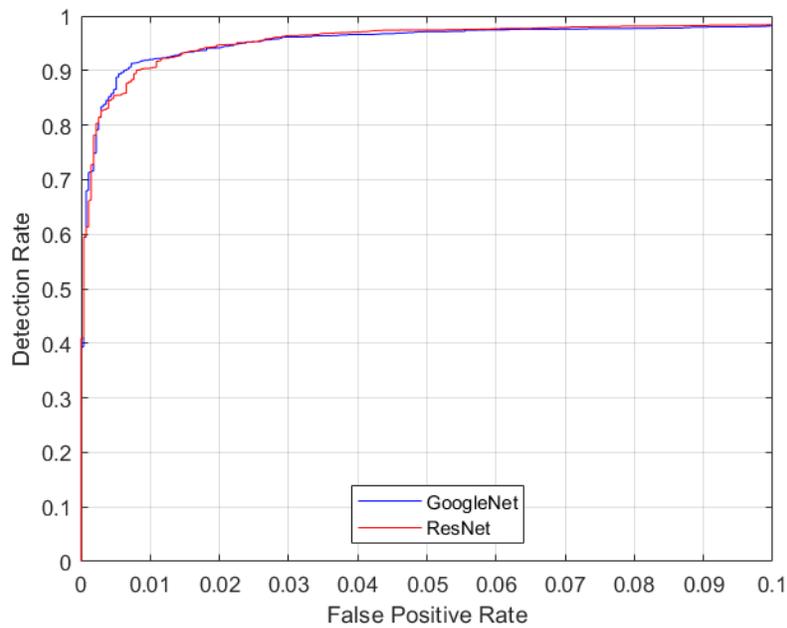

**Figure 7:** ROC curves for malaria detection.

## 2.4. Class Activation Mapping

In this section, we study the CAM results obtained using GoogLeNet and ResNet. Figures 8 and 9 present the CAM results obtained for two different cases from the malaria dataset using our proposed approach. Figure 8 presents the results for the case marked as parasitized by the expert reader, and our algorithm not only accurately predicts the same but also presents the discriminative region that contributed the most to its decision. The ROI is around the red spot containing plasmodium. Figure 9 presents the results for the case marked as uninfected. CAM visualization for 100 different cases using our presented approach is available at [12]. Typically, ResNet CAM converges to a smaller ROI in comparison to GoogLeNet. This type of automated CAD technology for malaria detection would assist the microscopists and enhance their workflow.

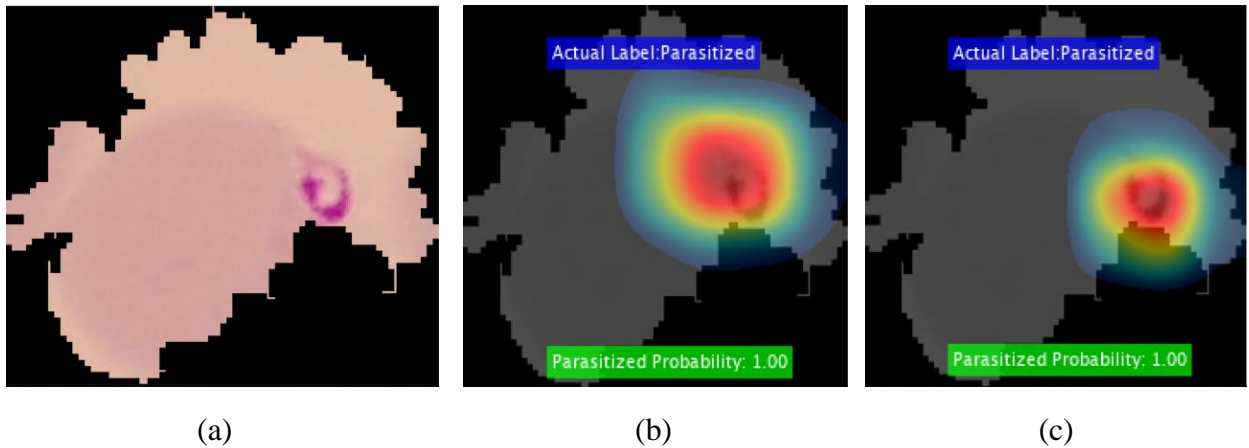

(a) (b) (c)

**Figure 8:** Typical CAD system output for a parasitized case: (a) input image, (b) GoogLeNet CAM, (c) ResNet CAM.

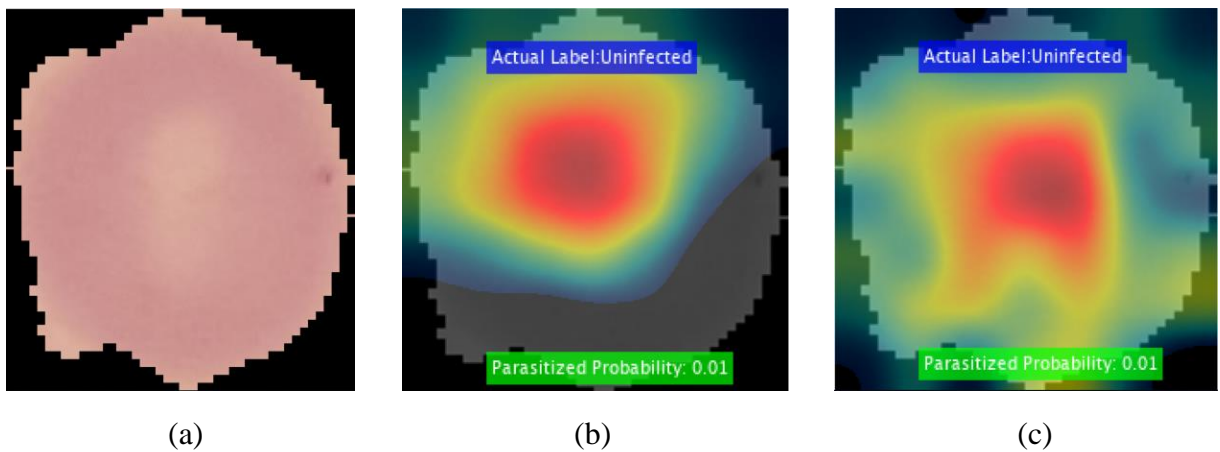

(a) (b) (c)

**Figure 9:** Typical CAD system output for an uninfected case: (a) input image, (b) GoogLeNet CAM, (c) ResNet CAM.

## 3. Diabetic Retinopathy Detection

In this section, we study the performance and visualize the CAM results of GoogLeNet and ResNet for the detection of DR on retinal images. DR is the leading cause of blindness affecting over 93 million people [13] across the world. An automated DR detection process would provide a valuable second opinion and quick solution in areas with a scarcity of trained clinicians.

### 3.1. Dataset

We make use of a publicly available dataset provided by the Asia Pacific Tele-Ophthalmology Society (APTOS) 2019 on Kaggle [14] to detect DR in retinal images. In this dataset, 3662 retinal images are graded by expert clinicians at Aravind Eye Hospital, India into 5 different categories: (i) Negative DR, (ii) Mild DR, (iii) Moderate DR, (iv) Proliferative DR, and (v) Severe DR. In this research, we solely focus on the detection of DR, hence, we merge all the mild, moderate, proliferative and severe cases into a single category 'positive DR'. Figures 10 and 11 present sample images marked in different categories by expert clinicians.

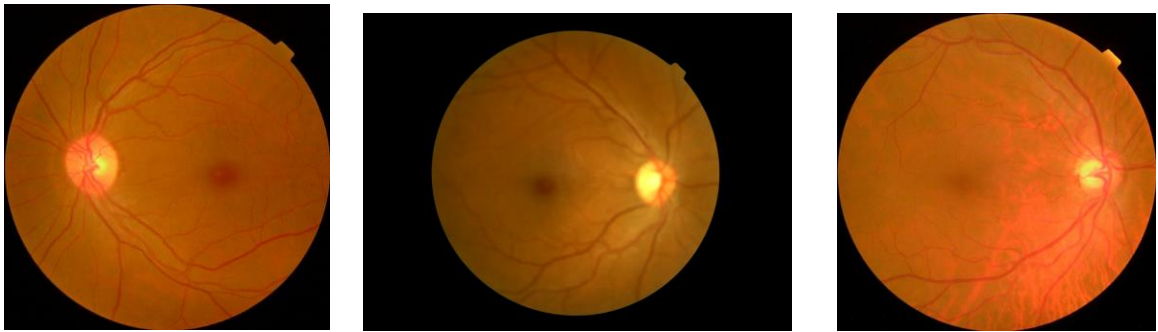

**Figure 10**: Sample retinal images marked as 'negative DR' by expert clinicians.

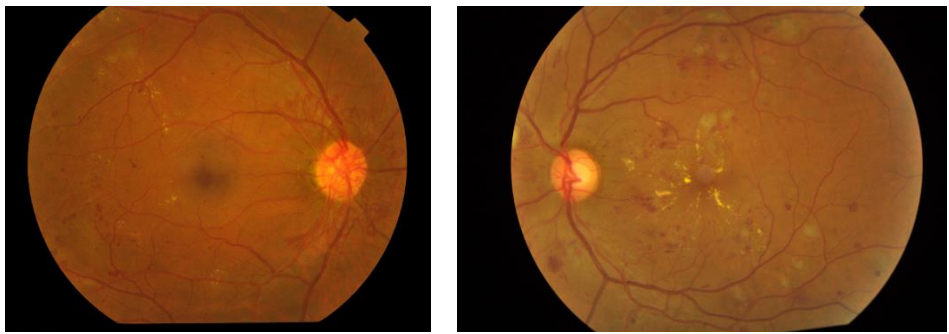

**Figure 11**: Sample retinal images marked as 'positive DR' by expert clinicians.

Similar to malaria detection (Section 2), we perform a hold-out validation study. We split the dataset into groups of 72%, 8%, and 20% for training, validation, and testing respectively. Table 3 presents the distribution of each dataset. There is no preprocessing except converting these images into the input size of the network.

Table 3: DR dataset distribution.

| Type of Dataset | # Positive DR images | # Negative DR images |
|---|---|---|
| Training | 1300 | 1337 |
| Validation | 144 | 149 |
| Testing | 361 | 371 |

## 3.2. Performance Summary

Aforementioned, we study the results obtained using the transfer learning based approaches with ResNet and GoogLeNet. Figures 12 and 13 present the confusion matrices and ROC curves obtained using these approaches for DR detection. Table 4 summarizes the performance in terms of overall accuracy and AUC for DR detection using these networks.

Table 4: Overall accuracy and AUC for DR detection.

| Method | Overall Accuracy (%) | AUC |
|---|---|---|
| GoogLeNet | 97.3 | 0.9943 |
| ResNet | 96.2 | 0.9939 |

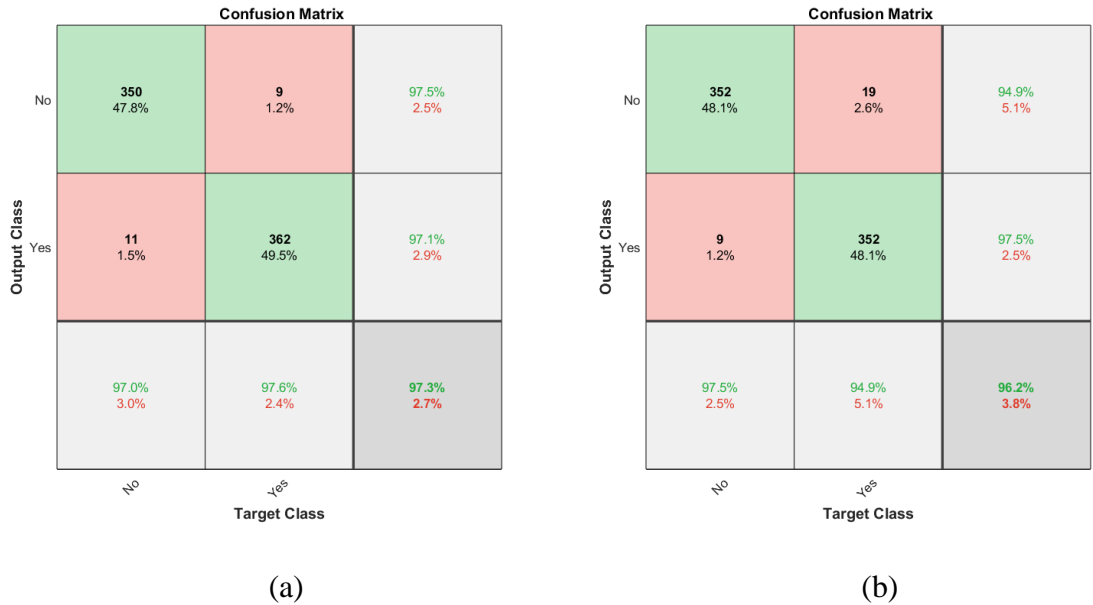

(a)        (b)

**Figure 12:** Confusion matrices obtained for DR detection: (a) GoogLeNet and (b) ResNet.

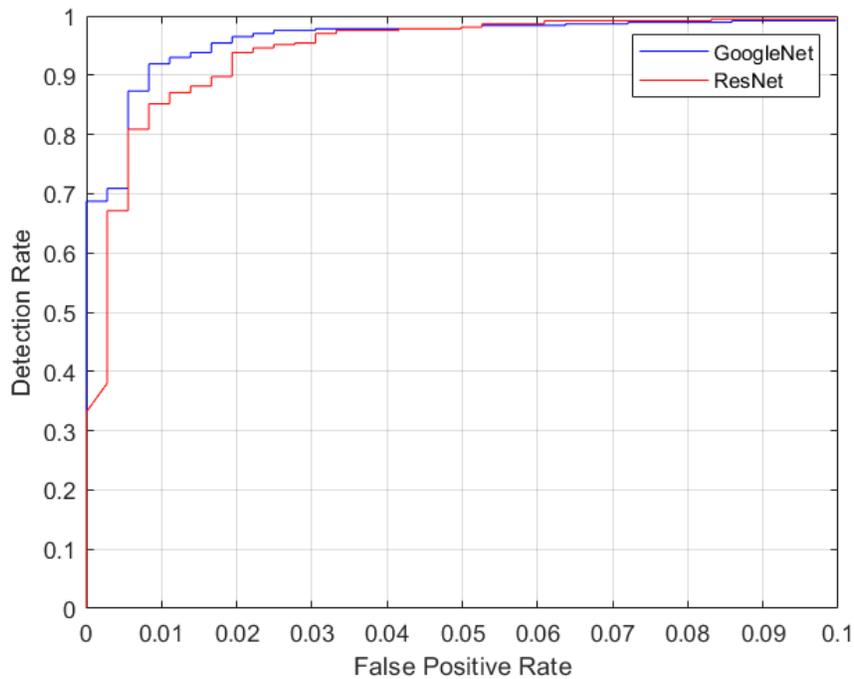

**Figure 13:** ROC curves for DR detection.

*3.3. Class Activation Mapping*

Figures 14 and 15 present the CAM results obtained for two different test cases from the APTOS dataset using our proposed approach. Figure 14 presents the results for a case marked as 'positive DR' by the expert clinicians. It is also interesting to note that CAM results presented by GoogLeNet and ResNet for DR detection have minimal intersection despite exhibiting similar performance, which could affect an expert clinician's choice of network. Figure 15 presents the results for a case marked as 'negative DR' by an expert clinician and the discriminative region for this case is near the retinal portion. CAM visualization for 60 different cases using our presented approach is available at [15]. This type of automated CAD technology for DR detection could be implemented for immediate solutions and could be applied in places with scarcity of such expert clinicians.

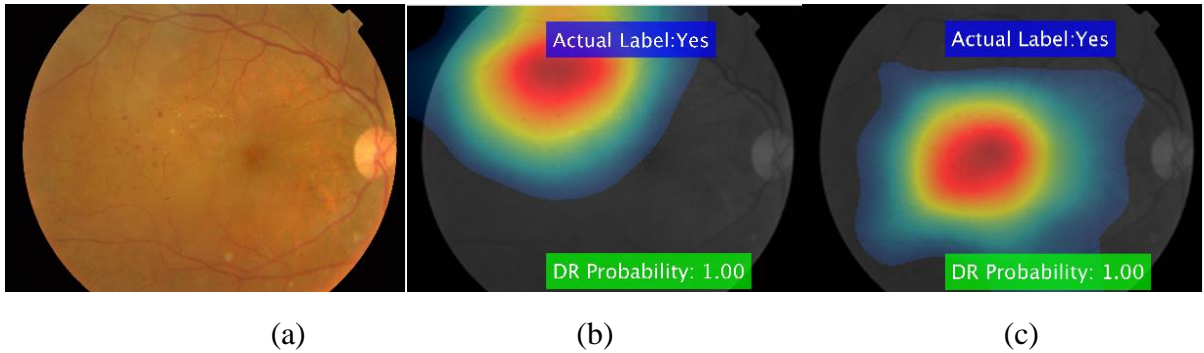

**Figure 14:** Typical CAD system output for a 'positive DR' case: (a) input image, (b) GoogLeNet CAM, (c) ResNet CAM.

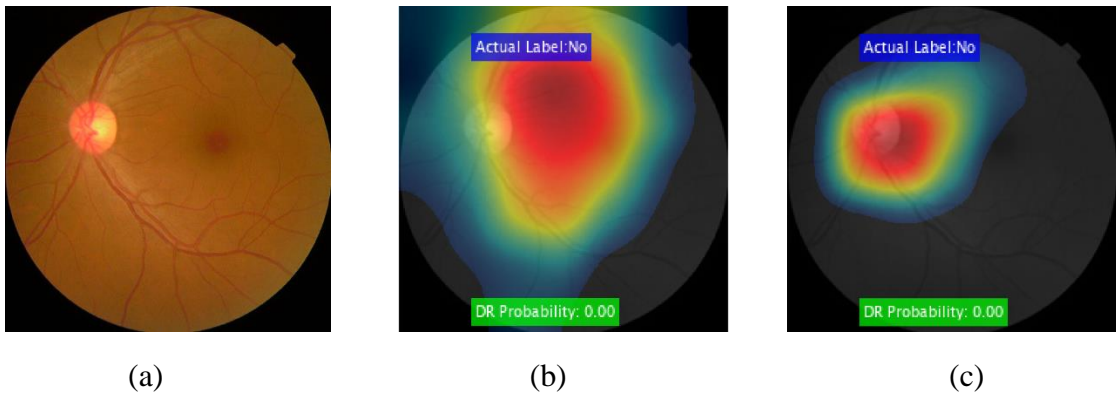

**Figure 15:** Typical CAD system output for a 'negative DR' case: (a) input image, (b) GoogLeNet CAM, (c) ResNet CAM.

## 4. Brain Tumor Detection

In this section, we study the performance and visualize the CAM results of GoogLeNet and ResNet for the detection of brain tumors on MRI scans.

### 4.1. Dataset

We make use of a publicly available dataset provided on Kaggle [16] for the classification of MRI scans. The dataset contains a total of 253 images classified into two different categories 'positive' and 'negative' brain tumor. Figures 16 and 17 present sample images marked as 'positive brain tumor' and 'negative brain tumor' by trained clinicians.

**Table 5**: Brain Tumor dataset distribution.

| Type of brain tumor | # MRI scans |
|---|---|
| Positive | 155 |
| Negative | 98 |

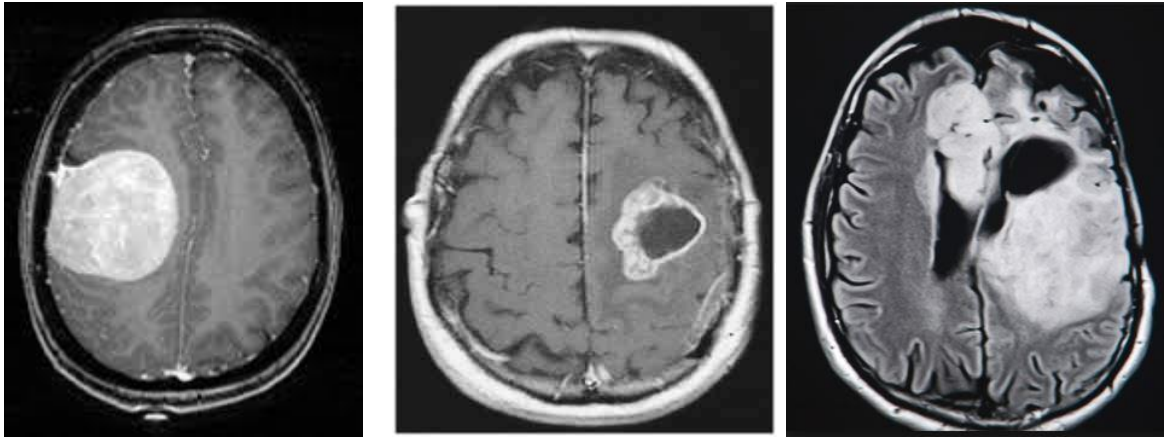

**Figure 16**: Sample MRI scans marked as 'positive brain tumor' by trained clinicians.

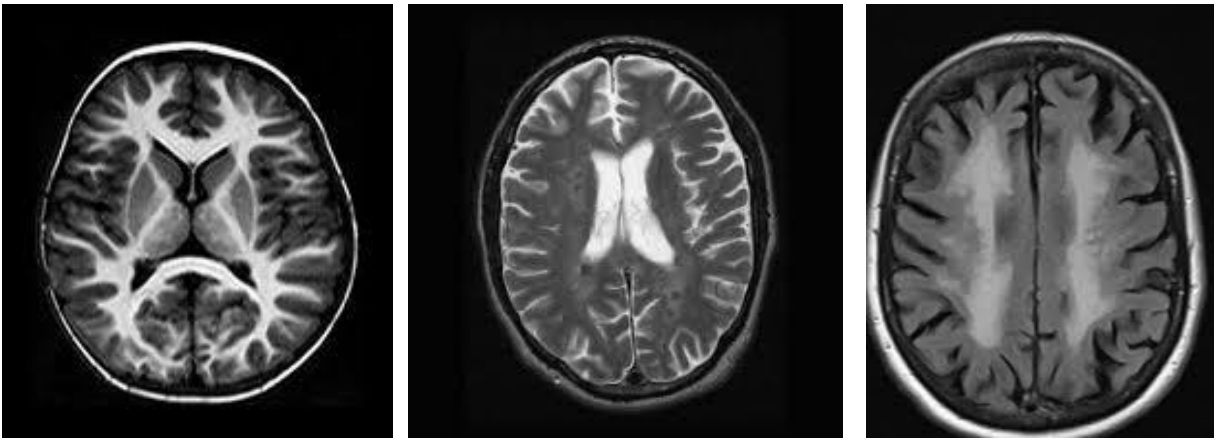

**Figure 17**: Sample MRI scans marked as 'negative brain tumor' by trained clinicians.

For this application, due to the limited availability of images, we perform a 10-fold validation study. We believe 10-fold validation would give a better estimate of our performance. *We train 10 different networks based on 9 folds and test on the remaining fold in each iteration*. For each fold, we train and tune our hyper-parameters solely based on the images from training fold. We make sure to exclude the testing fold in any manner to conduct a rigorous study. Note that we utilize the same set of cases in each fold for the architectures implemented. Overall distribution of the dataset is presented in Table 5.

*4.2. Preprocessing*

MRI scans contain text information for some cases, which are not essential for classification and might mislead our deep neural networks. Hence, we preprocess MRI scans by cropping ROI of brain and removing any additional text from the image using simple morphological operations. In addition, we perform histogram equalization to enhance and maintain the contrast across the dataset. Later, we resize the images to match with the input of ResNet and GoogLeNet architectures. Figure 18 presents the results obtained using these preprocessing techniques.

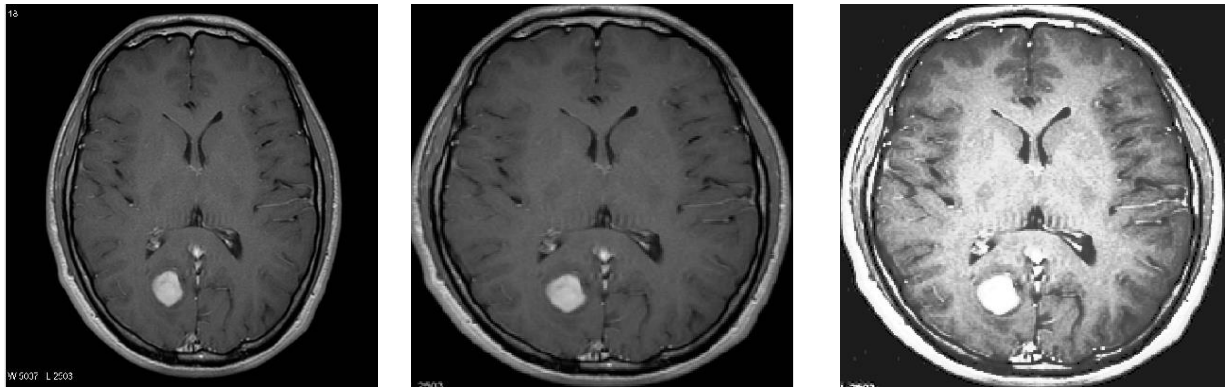

(a)             (b)             (c)

**Figure 18**: Preprocessing results: (a) input image, (b) image after cropping and resizing, (c) histogram equalization and resizing.

*4.3. Performance Summary*

Figures 19 and 20 present the confusion matrices and ROC curves obtained using these approaches for brain tumor detection. Table 6 presents the performance summary in terms of overall accuracy and AUC.

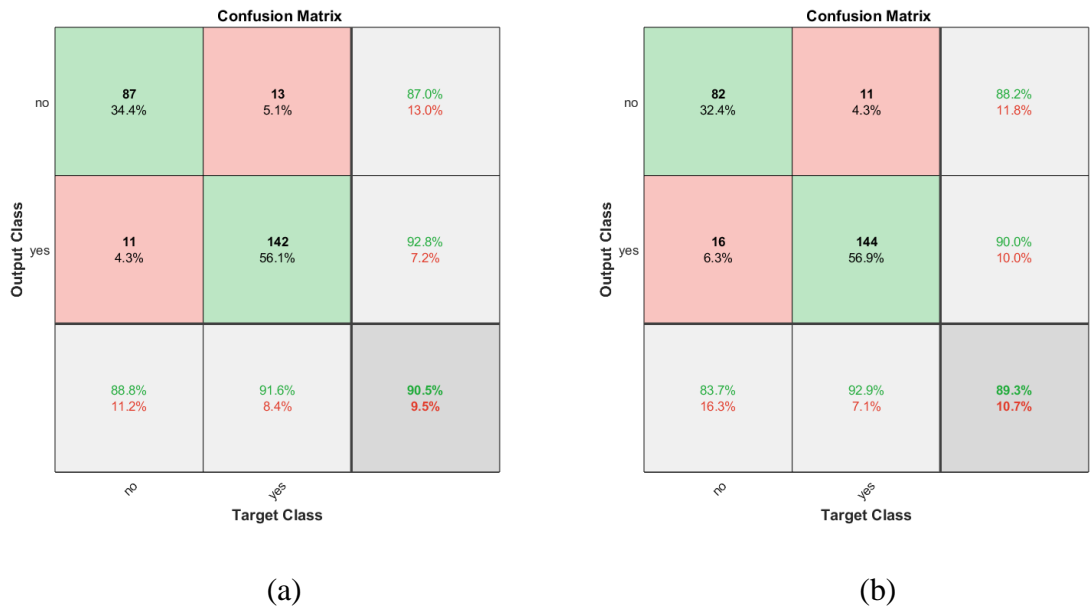

(a)             (b)

**Figure 19:** Confusion matrices obtained for brain tumor detection: (a) GoogLeNet and (b) ResNet.

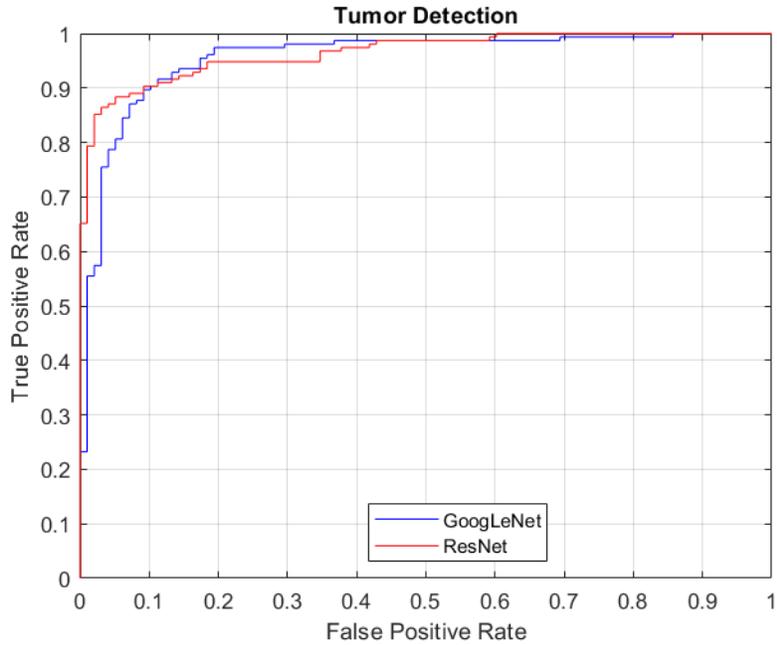

**Figure 20:** ROC curves for brain tumor detection.

**Table 6**: Overall accuracy and AUC for brain tumor detection.

| Method | Overall Accuracy (%) | AUC |
|---|---|---|
| GoogLeNet | 90.5 | 0.9559 |
| ResNet | 89.3 | 0.9650 |

*4.4. Class Activation Mapping*

Figures 21 and 22 present the CAM results obtained for two different cases from the brain tumor dataset. Figure 21 presents the results for the case marked as 'positive brain tumor' by the trained clinician and our algorithm accurately predicts the same. The discriminative region is near the tumor portion which would help the doctors pinpoint important features, spatially. Figure 22 presents the results for the case marked as 'negative brain tumor' and the visualization behind algorithm's prediction. CAM visualization for 26 (one test fold) different cases using our approach is available at [17]. This type of automated CAD technology for brain tumor detection would assist the doctors in providing a valuable second opinion and enhancing their workflow.

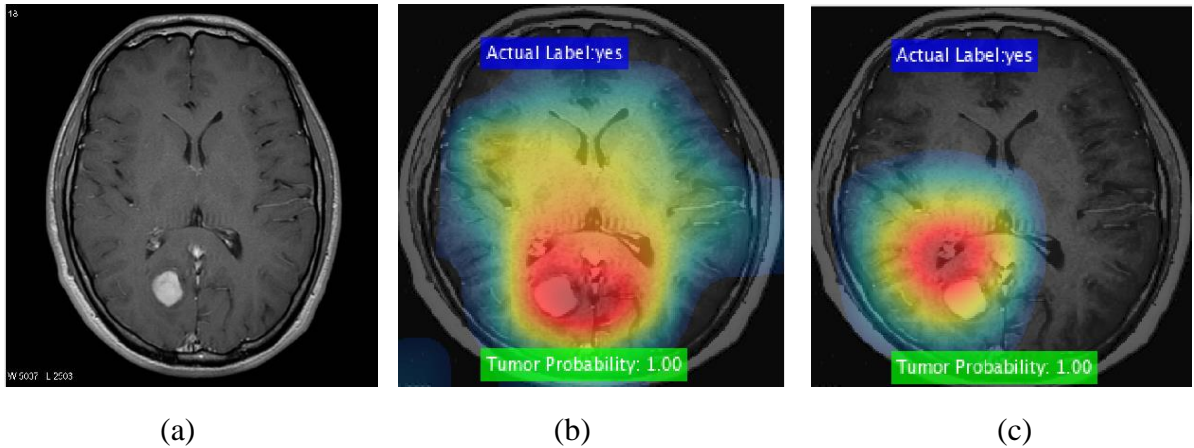

(a) (b) (c)

**Figure 21:** Typical CAD system output for a 'positive brain tumor' case: (a) input image, (b) GoogLeNet, (c) ResNet.

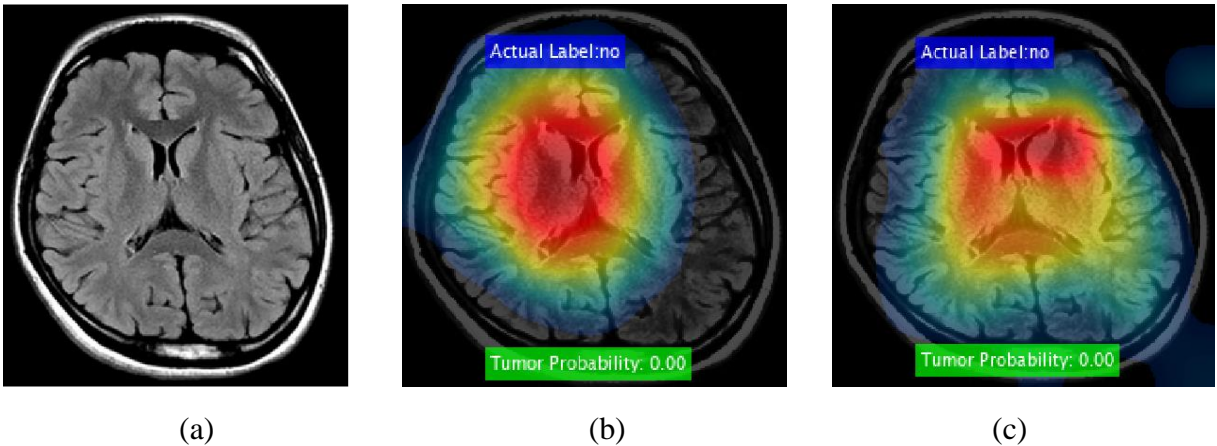

(a) (b) (c)

**Figure 22:** Typical CAD system output for a 'negative brain tumor' case: (a) input image, (b) GoogLeNet CAM, (c) ResNet CAM.

## 5. Tuberculosis Detection

In this section, we study the performance and visualize the CAM results of GoogLeNet and ResNet for the detection of tuberculosis on chest radiographs. Tuberculosis is created by mycobacterium and it severely affects the lung. According to World Health Organization (WHO), around 10 million new tuberculosis cases occurred in 2017 [18].

*5.1. Dataset*

We make use of the publicly available Shenzhen dataset [19] provided for the classification of chest radiographs. This dataset contains a total of 662 images. Figures 23 and 24 present sample images marked as 'Normal' and 'Tuberculosis' by radiologists.

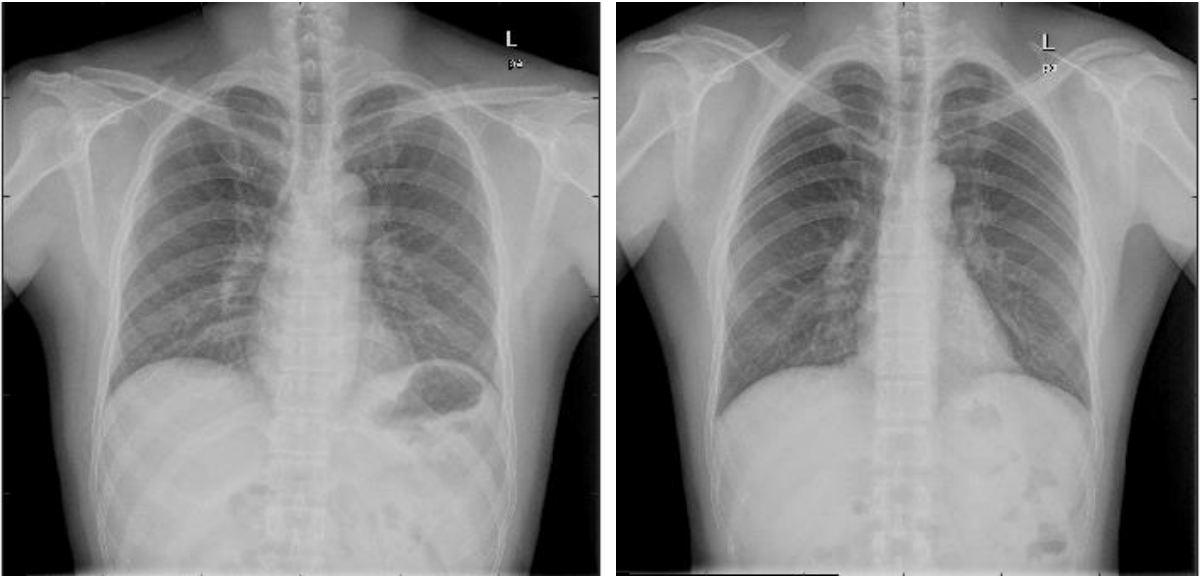

**Figure 23**: Sample chest radiographs marked as 'Normal' by radiologists

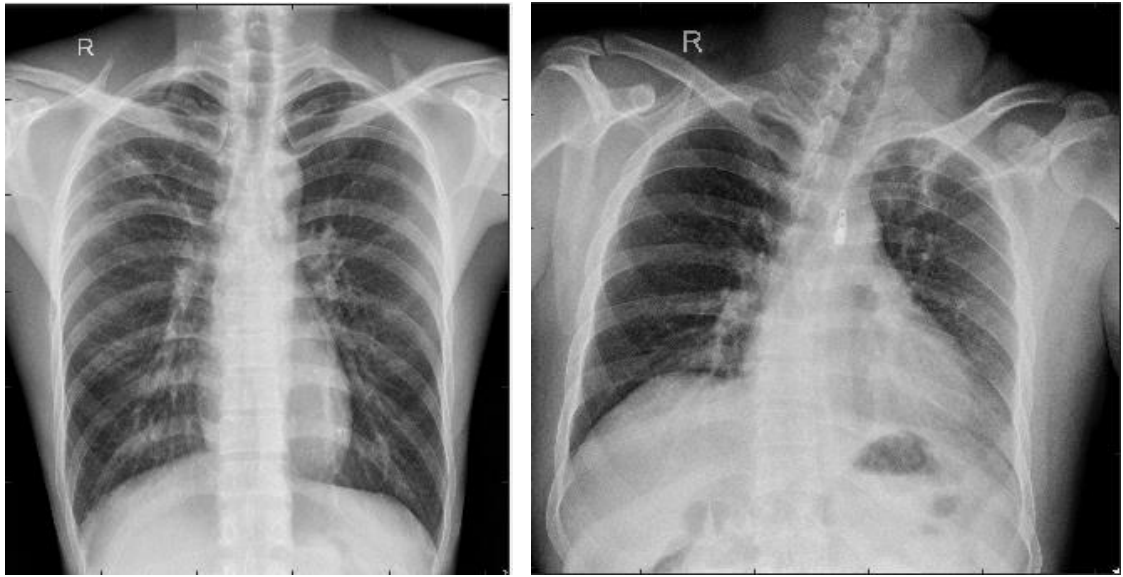

**Figure 24**: Sample chest radiographs marked as 'Tuberculosis' by radiologists

Similar to brain tumor detection, due to the limited availability of images, we perform a 10-fold validation study. Overall distribution of the dataset is presented in Table 7. There is no additional preprocessing except converting these images into the input size for the network.

**Table 7**: Tuberculosis dataset distribution

| Type | # Chest Radiographs |
|---|---|
| Normal | 326 |
| Tuberculosis | 336 |

## 5.2. Performance Summary

Figures 25 and 26 present the confusion matrices and ROC curves obtained using GoogLeNet and ResNet for tuberculosis detection. Table 8 presents the performance summary in terms of overall accuracy and AUC.

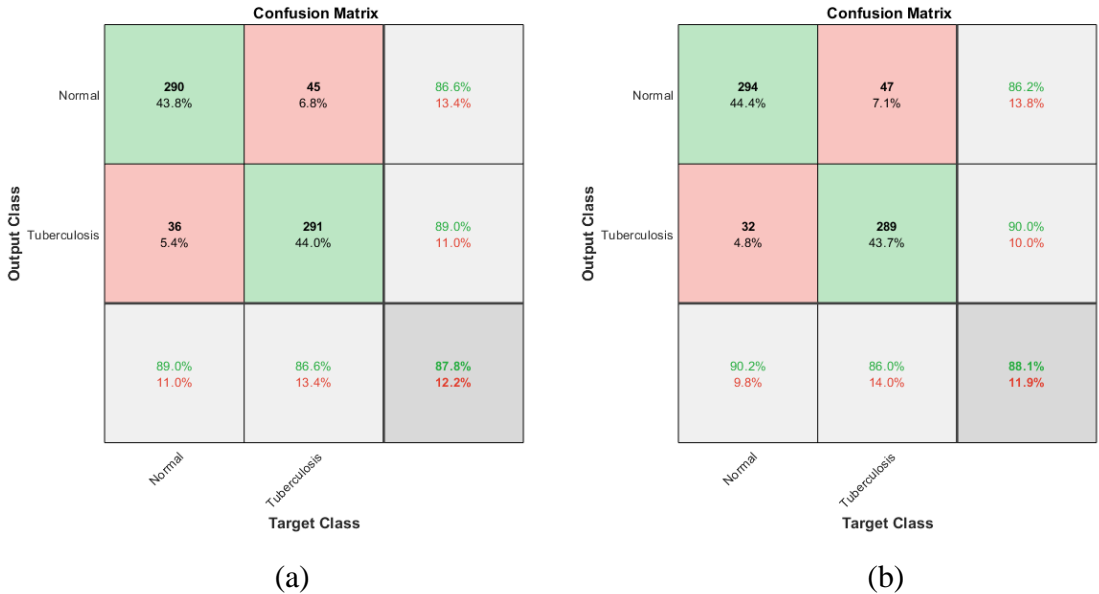

(a)          (b)

**Figure 25:** Confusion matrices obtained for tuberculosis detection: (a) GoogLeNet and (b) ResNet.

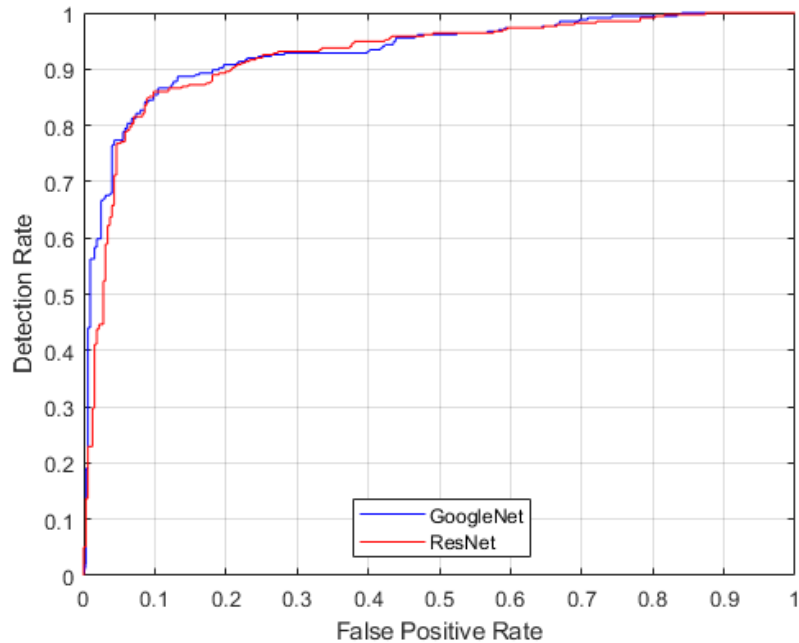

**Figure 26:** ROC curves for tuberculosis detection.

**Table 8**: Overall accuracy and AUC for tuberculosis detection.

| Method | Overall Accuracy (%) | AUC |
|---|---|---|
| GoogLeNet | 87.8 | 0.9317 |
| ResNet | 88.1 | 0.9245 |

*5.3. Class Activation Mapping*

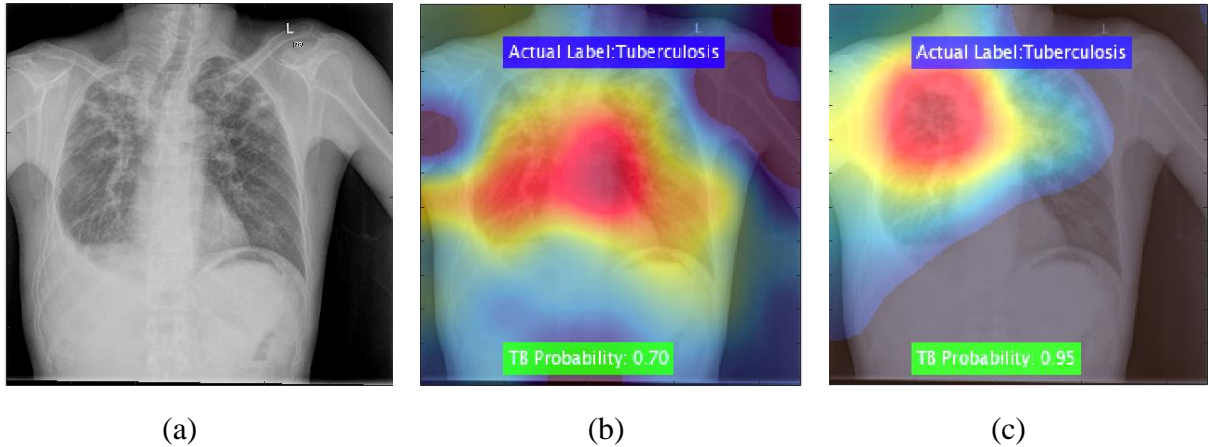

(a)        (b)        (c)

**Figure 27:** Typical CAD system output for a 'tuberculosis' case: (a) input image, (b) GoogLeNet, (c) ResNet.

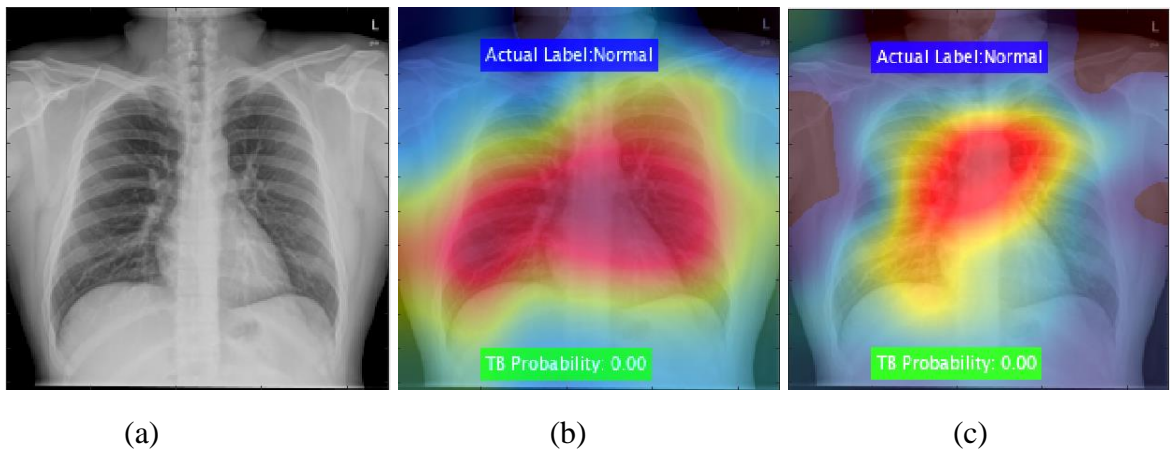

(a)        (b)        (c)

**Figure 28:** Typical CAD system output for a 'normal' case: (a) input image, (b) GoogLeNet, (c) ResNet.

Figures 27 and 28 present the results for tuberculosis detection in chest radiographs. Figure 27 presents the results for the case marked as 'tuberculosis' by the radiologist and our algorithm accurately predicts the same and also presents the discriminative region for its decision. It is interesting to note that GoogLeNet's CAM result indicates that the right lung of the patient is the discriminative region and provides a probability score of 0.95 whereas 'ResNet' believes in a different ROI and provides a probability score of 0.7. Figure 28 presents the results for a 'normal' case in chest radiographs and different regions are highlighted by both these algorithms for their prediction of the 'normal' category. Now, a radiologist can choose the network architecture based on these decisions and use them accordingly. CAM visualization results for certain random cases are available at [20]. This type of CAD technology would provide a valuable second opinion to the radiologists and enhance their understanding of machine learning models.

## 6. Discussions & Conclusions

In this research, we have presented results for various medical imaging applications that include CAD of malaria, diabetic retinopathy, brain tumors, and tuberculosis. Our proposed approach provides reasonable performance for all these applications. We have utilized the same classification architectures successfully for different imaging modalities that include microscopic images, retinal images, MRI scans, and chest radiographs, thereby demonstrating its efficacy. Table 9 summarizes the performance of our proposed algorithms for all applications in terms of overall accuracy.

**Table 9**: Performance summary in terms of overall accuracy (%) for different applications.

| CAD Application | GoogLeNet | ResNet |
|---|---|---|
| Malaria | 96.5 | **96.6** |
| Diabetic Retinopathy | **97.3** | 96.2 |
| Brain Tumor | **90.5** | 89.3 |
| Tuberculosis | 87.8 | **88.1** |

In addition, we have presented a comprehensive study of these algorithms based on their CAM results. This type of CAD system would let the medical expert analyst choose the algorithm based on their discretion in terms of CAM results, overall accuracy, AUC, or any other performance metric along with computation time and memory consumption. CAM results could be utilized by data science experts to further optimize their respective models in terms of architecture and/or preprocessing techniques. This type of CAM study would also assist the researchers in understanding the discriminative regions determined by various architectures in different imaging modalities. For instance, if an architecture achieves a high CAD detection accuracy for a particular dataset but the discriminative region determined using CAM is present in an unrelated area, performance metrics might mislead both data science researchers and medical expert analysts. Hence, CAM visualization would provide the necessary documentation for the medical expert analysts to enhance the trust in CAD system. This research can be further extended by studying the ROI determined by CAM by changing different hyper-parameters and how they evolve throughout the epochs. Another way to extend this research is by fusing CAM regions determined using different architectures.